%% file: dipole.tex
\documentclass[sigconf]{acmart}

\usepackage{booktabs} 
\usepackage{booktabs}
\usepackage{color}
\usepackage{bm}
\usepackage{algorithm}
\usepackage{algorithmic}
\usepackage{multirow}
\usepackage{graphicx}
\usepackage{threeparttable}
\usepackage{url}
\usepackage{subfigure}
\usepackage{here}

\usepackage{amsmath}
\usepackage{amssymb}
\usepackage{longtable}
 \usepackage{rotating}

 \usepackage{tabularx}
 \usepackage{array}

\setcopyright{rightsretained}

\acmDOI{10.475/123_4}

\acmISBN{123-4567-24-567/08/06}

\acmConference[KDD'17]{the 22nd ACM SIGKDD International Conference on Knowledge Discovery and Data Mining}{August 2017}{Halifax, NS, Canada}
\acmYear{2017}
\copyrightyear{2017}

\acmPrice{15.00}

\begin{document}

\title[{\sf Dipole}: Diagnosis Prediction in Healthcare via Attention-based BRNNs]
{{\sf Dipole}: Diagnosis Prediction in Healthcare via Attention-based Bidirectional Recurrent Neural Networks}

\author{Fenglong Ma}
\authornote{This work was mostly done when the first author was an intern in Xerox.}
\affiliation{%
  \institution{SUNY Buffalo}
  \city{Buffalo}
  \state{NY}
  \country{USA}
}
\email{fenglong@buffalo.edu}

\author{Radha Chitta}
\affiliation{%
  \institution{Conduent Labs US}
  \city{Rochester}
  \state{NY}
  \country{USA}
}
\email{radha.chitta@conduent.com}

\author{Jing Zhou}
\affiliation{%
  \institution{Conduent Labs US}
  \city{Rochester}
  \state{NY}
  \country{USA}
 }
\email{jing.zhou@conduent.com}

\author{Quanzeng You}
\affiliation{%
  \institution{University of Rochester}
  \city{Rochester}
  \state{NY}
  \country{USA}
 }
\email{quanzeng.you@rochester.edu}

\author{Tong Sun}
\authornote{This work was done when the fifth author was part of Xerox.}
\affiliation{%
  \institution{United Technologies Research Center}
  \city{East Hartford}
  \state{CT}
  \country{USA}
 }
\email{sunt@utrc.utc.com}

\author{Jing Gao}
\affiliation{%
  \institution{SUNY Buffalo}
  \city{Buffalo}
  \state{NY}
  \country{USA}
}
\email{jing@buffalo.edu}

\renewcommand{\shortauthors}{F. Ma et. al.}

\input{0-abstract}

%
%


\copyrightyear{2017}
\acmYear{2017}
\setcopyright{acmcopyright}
\acmConference{KDD'17}{}{August 13--17, 2017, Halifax, NS, Canada.}
\acmPrice{15.00}
\acmDOI{http://dx.doi.org/10.1145/3097983.3098088}
\acmISBN{978-1-4503-4887-4/17/08}

 \fancyhead{}


\maketitle

\input{1-introduction}
\input{6-relatedwork}

\input{3-model}
\input{5-experiments}

\input{7-conclusions}

\begin{acks}
  The authors would like to thank the anonymous referees for
  their valuable comments and helpful suggestions. 
  This work is supported in part by the US National Science
Foundation under grants IIS-1319973, IIS-1553411 and IIS-1514204.
Any opinions, findings, and conclusions or recommendations
expressed in this material are those of the author(s) and do not
necessarily reflect the views of the National Science Foundation.
\end{acks}

\bibliographystyle{ACM-Reference-Format}
\bibliography{kdd}
\end{document}

%% file: 0-abstract.tex
\begin{abstract}
Predicting the future health information of patients from the historical Electronic Health Records (EHR) is a core research task in the development of personalized healthcare.
Patient EHR data consist of sequences of visits over time, where each visit contains multiple medical codes, including diagnosis, medication, and procedure codes.
The most important challenges for this task are to model the temporality and high dimensionality of sequential EHR data and to interpret the prediction results.
Existing work solves this problem by employing recurrent neural networks (RNNs) to model EHR data and utilizing simple attention mechanism to interpret the results.
However, RNN-based approaches suffer from the problem that the performance of RNNs drops when the length of sequences is large, and the relationships between subsequent visits are ignored by current RNN-based approaches.
To address these issues, we propose {\sf Dipole}, an end-to-end, simple and robust model for predicting patients' future health information. 
 {\sf Dipole} employs bidirectional recurrent neural networks to remember all the information of both the past visits and the future visits, and it introduces three attention mechanisms to measure the relationships of different visits for the prediction.
 With the attention mechanisms,  {\sf Dipole} can interpret the prediction results effectively. 
 {\sf Dipole} also allows us to interpret the learned medical code representations which are confirmed positively by medical experts.
 Experimental results on two real world EHR datasets show that the proposed {\sf Dipole} can significantly improve the prediction accuracy compared with the state-of-the-art diagnosis prediction approaches and provide clinically meaningful interpretation.
\end{abstract}

%% file: 1-introduction.tex
\section{Introduction}
\emph{Electronic Health Records} (EHR), consisting of longitudinal patient health data, including demographics, diagnoses, procedures, and medications, have been utilized successfully in several predictive modeling tasks in healthcare~\cite{choi2016retain, choi2016gram, choi2016mlr, zhou2013patient}. 
EHR data are temporally sequenced by patient medical visits that are represented by a set of high dimensional clinical variables (i.e., medical codes). 
One critical task is to predict the future diagnoses based on patient's historical EHR data, i.e., \emph{diagnosis prediction}. 
When predicting diagnoses, each patient's visit and the medical codes in each visit may have varying importance. 
Thus, the most important and challenging issues in diagnosis prediction are:
\begin{itemize}
\item How to correctly model such \emph{temporal} and \emph{high dimensional} EHR data to significantly improve the performance of prediction; 
\item How to reasonably \emph{interpret} the importance of visits and medical codes in the prediction results.
\end{itemize}

In order to model sequential EHR data, recurrent neural networks (RNNs) have been employed in the literature for deriving accurate and robust representations of patient visits in diagnosis prediction task~\cite{choi2016retain, choi2016gram}. 
{\sf RETAIN}~\cite{choi2016retain} and {\sf GRAM}~\cite{choi2016gram} are two state-of-the-art models utilizing RNNs for predicting the future diagnoses. 
{\sf RETAIN} applies an RNN with reverse time ordered EHR sequences, while {\sf GRAM} uses an RNN when modeling time ordered patient visits. 
Both models achieve good prediction accuracy. 
However, they are constrained by the forgetfulness associated with models using RNNs, i.e. RNNs cannot handle long sequences effectively. 
The predictive power of these models drops significantly when the length of the patient visit sequences is large. 
Bidirectional recurrent neural networks (BRNNs)~\cite{schuster1997bidirectional}, which can be trained using all available input information in the past and future, have been used to alleviate the effect of the long sequence problem, and improve the predictive performance. 

However, it is infeasible to interpret the outputs of models incorporating either RNNs or BRNNs. 
Interpretability is crucial in the healthcare domain, as it can lead to the identification of potential risk factors and the design of suitable intervention mechanisms. 
Non-temporal models such as Med2Vec~\cite{choi2016mlr} generate easily interpretable low-dimensional representations of the medical codes, but do not account for the temporal nature of the EHR data. 

To model the temporal EHR data and interpret the prediction results simultaneously, attention-based neural networks can be applied, which aim to learn the relevance of the data samples to the task.  For example, {\sf RETAIN}~\cite{choi2016retain} employs location-based attention to predict the future diagnosis. It calculates the attention weights for a visit at time $t$, using the medical information in the current visit and the hidden state of the recurrent neural network at time $t$, to predict the visit at time $t+1$. It ignores the relationships between all the visits from time $1$ to time $t$. We believe that accounting for all the past visit information may help the predictive models to improve the accuracy and provide better interpretation.

To tackle all the aforementioned issues and challenges, we propose an efficient and accurate \underline{\textbf{di}}agnosis \underline{\textbf{p}}rediction m\underline{\textbf{o}}d\underline{\textbf{el}} ({\sf Dipole}) using attention-based bidirectional recurrent neural networks for learning low-dimensional representations of the patient visits, and employ the learned representations for future diagnosis prediction. The learned representations are easily interpretable, and can also be used to learn how important each visit is to the future diagnosis prediction. Specifically, it first embeds the high dimensional medical codes (i.e., clinical variables) into a low code-level space, and then feeds the code representations into an attention-based bidirectional recurrent neural network to generate the hidden state representation. The hidden representation is fed through a softmax layer to predict the medical codes in future visits. We experiment with three types of attention mechansims: (i) location-based, (ii) general, and (iii) concatenation-based, to calculate the attention weights for all the prior visits for each patient. These mechanisms model the inter-visit relationships, where the attention weights represent the importance of each visit.

We demonstrate that the proposed {\sf Dipole} achieves significantly higher prediction accuracy when compared to the state-of-the-art approaches in diagnosis prediction, using two datasets derived from Medicaid claims data.
A case study is conducted to show that the proposed model accurately assigns varying attention weights to past visits.
We evaluate the interpretability of the learned representations through qualitative analysis. 
Finally, we illustrate the reasonableness of employing bidirectional recurrent neural networks to model temporal patient visits. 
In summary, our main contributions are as follows:
\begin{itemize}
\item We propose {\sf Dipole}, an end-to-end, simple and robust model to accurately predict the future visit information and reasonably interpret the prediction results, without depending on any expert medical knowledge.
\item {\sf Dipole} models patient visit information in a time-ordered and reverse time-ordered way and employs three attention mechanisms to calculate the weights for previous visits.
\item We empirically show that the proposed {\sf Dipole} outperforms existing methods in diagnosis prediction on two large real world EHR datasets.
\item We analyze the experimental results with clinical experts to validate the interpretability of the learned medical code representations.
\end{itemize}

The rest of this paper is organized as follows: 
In Section \ref{relatedwork}, we discuss the connection of the proposed approaches to related work. Section \ref{model} presents the details of the proposed {\sf Dipole}. The experimental results are presented in Section \ref{experiments}. Section \ref{end} concludes the paper.

%% file: 6-relatedwork.tex
\section{Related Work}\label{relatedwork}
This section reviews the existing work for mining Electronic Healthcare Records (EHR) data. In particular, we focus on the state-of-the-art models on diagnosis prediction task. We also introduce some work using attention mechanisms.

\subsection{EHR Data Mining}
Mining EHR data is a hot research topic in healthcare informatics. The tackled tasks include electronic genotyping and phenotyping~\cite{jensen2012mining, liu2015temporal, che2015deep, zhou2014micro}, disease progression~\cite{wang2014unsupervised,choi2015constructing, zhou2011multi,xiao2017learning}, adverse drug event detection~\cite{ma2017unsupervised}, diagnosis prediction~\cite{choi2016retain, choi2016gram, choi2016mlr, zhou2013patient, suo2017amia}, and so on. In most tasks, deep learning models can significantly improve the performance. Recurrent neural networks (RNNs) can be used for modeling multivariate time series data in healthcare with missing values~\cite{che2016recurrent, lipton2016modeling}. Convolutional neural networks (CNNs) are used to predict unplanned readmission~\cite{nguyen2016deepr} and risk~\cite{cheng2016risk} with EHR. Stacked denoising autoencoders (SDAs) are employed to detect the characteristic patterns of physiology in clinical time series data~\cite{che2015deep}.

Diagnosis prediction is an important and difficult task in healthcare. {\sf Med2Vec}~\cite{choi2016mlr} aims to learn the representations of medical codes, which can be used to predict the future visit information. This method ignores long-term dependencies of medical codes among visits. {\sf RETAIN}~\cite{choi2016retain} is an interpretable predictive model, which employs reverse time attention mechanism in an RNN for binary prediction task. {\sf GRAM}~\cite{choi2016gram} is a graph-based attention model for healthcare representation learning, which uses medical ontologies to learn robust representations and an RNN to model patient visits. Both {\sf RETAIN} and {\sf GRAM} apply attention mechanisms and improve the prediction performance.

\medskip
Compared with the aforementioned predictive models, the proposed approaches not only employ bidirectional neural networks when modeling visit information but also design different attention mechanisms to assign different weights for the past visits. 
Relying on these two properties, the proposed {\sf Dipole} can improve the prediction performance significantly and interpret the meanings of medical codes reasonably.

\subsection{Attention-based Neural Networks}
Attention-based neural networks have been successfully used in many tasks~\cite{ba2015multiple, xu2015show, you2016image, luong2015effective, bahdanau2015neural, hermann2015teaching, chorowski2015attention,lamb2016professor}. 
Specifically for neural machine translation~\cite{luong2015effective, bahdanau2015neural}, given a sentence in the original language (i.e., original space), RNNs were adopted to generate the word representations in the sentence $\bm{h}_1, \cdots, \bm{h}_{|S|}$, where $|S|$ is the number of words in this sentence.
In order to find the $t$-th word in the target language (or target space), a weight $\alpha_{ti}$, i.e., attention score, is assigned to each word in the original language. Then, a context vector $\bm{c}_t = \sum_{i=1}^{|S|} \alpha_{ti} \bm{h}_i$ is calculated to predict the $t$-th word in the target language. However, diagnosis prediction task is different from the language translation task as all the visits for each patient are in the same space.

%% file: 3-model.tex
\section{Methodology}\label{model}
In this section, we first introduce the structure of EHR data and some basic notations. Then we describe the details of the proposed {\sf Dipole} neural network. Finally, we describe the interpretation for the learned code representations and visit representations.

\subsection{Basic Notations}
We denote all the unique medical codes from the EHR data as $c_1, c_2, \cdots, c_{|\mathcal{C}|} \in \mathcal{C}$, where $|\mathcal{C}|$ is the number of unique medical codes. Assuming there are $N$ patients, the $n$-th patient has $T^{(n)}$ visit records in the EHR data. The patient can be represented by a sequence of visits $V_1, V_2, \cdots, V_{T^{(n)}}$. Each visit $V_t$, containing a subset of medical codes ($V_t \subseteq \mathcal{C}$), is denoted by a binary vector $\bm{x}_t \in \{0, 1\}^{|\mathcal{C}|}$, where the $i$-th element is 1 if $V_t$ contains the code $c_i$. 
Each diagnosis code can be mapped to a node of the International Classification of Diseases (ICD-9)\footnote{\url{https://en.wikipedia.org/wiki/List_of_ICD-9_codes}}, and each procedure code can be mapped to a node in the Current Procedural Terminology (CPT)\footnote{\url{https://en.wikipedia.org/wiki/Current_Procedural_Terminology}}. 
Below we use a simple example to illustrate the problem.

There are two diagnosis codes: {250} (\emph{Diabetes mellitus}) and {254} (\emph{Diseases of thymus gland}), and one procedure code {11720} (\emph{Debride nail, 1-5}) in the whole dataset, i.e., ${|\mathcal{C}|} = 3$. 
If the medical codes in the $t$-th patient visit are {250} and {254}, then $\bm{x}_t = [1, 1, 0]$.

Both ICD-9 and CPT systems are coded hierarchically, which means that each medical code has a ``parent'', i.e., category label. For example, the diagnosis codes {250} and {254} belong to the same category \emph{Diseases of other endocrine glands}, and the procedure code {11720} is in the category \emph{Surgical procedures on the nails}. Thus, each visit $V_t$ has a corresponding coarse-grained category representation $\bm{y}_t \in \{0, 1\}^{|\mathcal{G}|}$, where $|\mathcal{G}|$ is the unique number of categories. In the above example, $|\mathcal{G}| = 2$, and $\bm{y}_t = [1, 0]$. For simplicity, we describe the proposed algorithm for a single patient and drop the superscript $(n)$ when it is unambiguous. The input of the proposed {\sf Dipole} model is a time-ordered sequence of patient visits.

\subsection{Model}\label{sub_model}
\begin{figure}[!htb]
 \centering
  \includegraphics[width=2.8in]{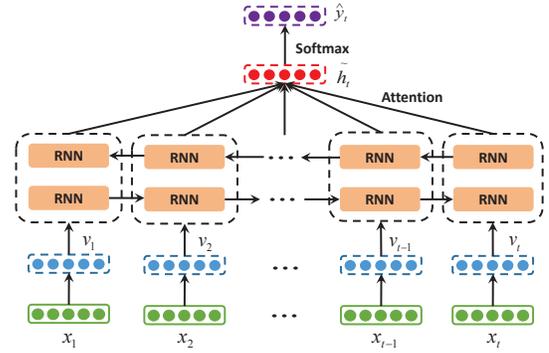}
 \caption{The Proposed {\sf Dipole} Model.}\label{fig:dipole}
\end{figure}

The goal of the proposed algorithm is to predict the $(t+1)$-th visit's category-level medical codes. 
Figure \ref{fig:dipole} shows the high-level overview of the proposed model. 
Given the visit information from time $1$ to $t$, the $i$-th visit's medical codes $\bm{x}_i$ can be embedded into a vector representation $\bm{v}_i$. 
The vector $\bm{v}_i$ is fed into the Bidirectional Recurrent Neural Network (BRNN)~\cite{schuster1997bidirectional}, which outputs a hidden state $\bm{h}_i$ as the representation of the $i$-th visit.
Along with the set of hidden states $\{\bm{h}_i\}_{i=1}^{t-1}$, we are able to compute the relative importance vector $\bm{\alpha}_t$ for the current visit $t$.
Subsequently, a \emph{context} state $\bm{c}_{t}$ is computed from the relative importance $\bm{\alpha}_t$ and $\{\bm{h}_i\}_{i=1}^{t-1}$. 
This procedure is known as attention model~\cite{bahdanau2015neural}, which will be detailed in the following sections. 
Next, from the context state $\bm{c}_t$ and the current hidden state $\bm{h}_t$, we can obtain an \emph{attentional hidden} state $\tilde{\bm{h}}_t$, which is used to predict the category-level medical codes appearing in the $(t+1)$-th visit, i.e., $\hat{\bm{y}}_{t}$. 
The proposed neural network can be trained end-to-end.

\smallskip\noindent\textbf{Visit Embedding}\\
Given a visit $\bm{x}_i \in \{0,1\}^{|\mathcal{C}|}$, we can obtain its vector representation $\bm{v}_i \in \mathbb{R}^m$ as follows:
\begin{equation}\label{eq:visit}
\bm{v}_i = \text{ReLU}(\bm{W}_v \bm{x}_i + \bm{b}_c),
\end{equation}
where $m$ is the size of embedding dimension, $\bm{W}_v \in \mathbb{R}^{m \times |\mathcal{C}|}$ is the weight matrix of medical codes, and $\bm{b}_c \in \mathbb{R}^m$ is the bias vector. $\text{ReLU}$ is the rectified linear unit defined as $\text{ReLU}(\bm{v}) = \max(\bm{v, 0})$, where $\max()$ applies element-wise to vectors. The reason we employ the rectified linear unit as the activation function is that $\text{ReLU}$ enables the learned vector representations to be interpretable~\cite{choi2016mlr}.

\medskip\noindent\textbf{Bidirectional Recurrent Neural Networks}\\
Recurrent Neural Networks (RNNs) provide a very elegant way of modeling sequential healthcare data \cite{choi2016retain,choi2016gram}. However, one drawback of RNNs is that the prediction performance will drop when the length of the sequence is very large \cite{schuster1997bidirectional}. In order to overcome this drawback, we employ Bidirectional Recurrent Neural Networks (BRNNs) in the proposed model which can be trained using all the available input visits' information from two directions to improve the prediction performance. Note that we use ``RNNs'' to denote any Recurrent Neural Networks variant dealing with the vanishing gradient problem~\cite{bengio1994learning}, such as Long-Short Term Memory (LSTM)~\cite{hochreiter1997long} and Gated Recurrent Unit (GRU)~\cite{cho2014properties}. In our implementation, we use GRU to adaptively capture dependencies among patient visit information.

A BRNN consists of a forward and backward RNN. The forward RNN $\overrightarrow{f}$ reads the input visit sequence from $\bm{x}_1$ to $\bm{x}_T$ and calculates a sequence of \emph{forward hidden states} $(\overrightarrow{\bm{h}}_1, \cdots, \overrightarrow{\bm{h}}_T)$ ($\overrightarrow{\bm{h}}_i \in \mathbb{R}^p$ and $p$ is the dimensionality of hidden states). The backward RNN $\overleftarrow{f}$ reads the visit sequence in the reverse order, i.e., from $\bm{x}_T$ to $\bm{x}_1$, resulting in a sequence of \emph{backward hidden states} $(\overleftarrow{\bm{h}}_1, \cdots, \overleftarrow{\bm{h}}_T)$ ($\overleftarrow{\bm{h}}_i \in \mathbb{R}^p$). By concatenating the forward hidden state $\overrightarrow{\bm{h}}_i$ and the backward one $\overleftarrow{\bm{h}}_i$, we can obtain the final latent vector representation as $\bm{h}_i = [\overrightarrow{\bm{h}}_i; \overleftarrow{\bm{h}}_i]^{\top}$ ($\bm{h}_i \in \mathbb{R}^{2p}$).
Note that the future visit information is only used when training the model. Only the past visit information is provided to predict the future visits during the testing phase.

\medskip\noindent\textbf{Attention Mechanism}\\
In diagnosis prediction task, the final goal is to predict the category-level medical codes of the $(t+1)$-th visit, i.e., $\bm{y}_{t}$, according to the visits from $\bm{x}_1$ to $\bm{x}_t$. The output of the $t$-th visit $\bm{x}_t$ ($\bm{h}_t$) is the estimated vector representation of the $(t+1)$-th visit. However, it may contain partial visit information to be predicted. Thus, how to derive a context vector $\bm{c}_t$ that captures relevant information to help predict the future visit $\bm{y}_{t}$ is the key issue. There are three methods that can be used to compute the context vector $\bm{c}_t$:

\smallskip
$\bullet$ \emph{Location-based Attention}.
A location-based attention function is to calculate the weights solely from the current hidden state $\bm{h}_i$ as follows:
\begin{equation}\label{location}
\alpha_{ti} = \bm{W}_{\alpha}^{\top} \bm{h}_i + b_{\alpha},
\end{equation}
where $\bm{W}_{\alpha} \in \mathbb{R}^{2p}$ and $b_{\alpha} \in \mathbb{R}$ are the parameters to be learned. 
According to Eq.~(\ref{location}), we can obtain an attention weight vector $\bm{\alpha}_t$ using softmax function as follows:
\begin{equation}
\label{local_attention}
\bm{\alpha}_t = \text{Softmax} ([\alpha_{t1}, \alpha_{t2}, \cdots, \alpha_{t(t-1)}]).
\end{equation}
Then the context vector $\bm{c}_t  \in \mathbb{R}^{2p}$ can be calculated based on the weights obtained from Eq.~(\ref{local_attention}) and the hidden states from $\bm{h}_1$ to $\bm{h}_{t-1}$ as follows:
\begin{equation}\label{ct}
\bm{c}_t = \sum_{i = 1}^{t-1} \alpha_{ti} \bm{h_i}.
\end{equation}

Since location-based attention mechanism only considers each individual hidden state information, it does not capture the relationships between the current hidden state and all the previous hidden states. To utilize the information from all the previous hidden states, we adopt the following two attention mechanisms in the proposed {\sf Dipole} .

\smallskip$\bullet$ \emph{General Attention}.
An easy way to capture the relationship between $\bm{h}_t$ and $\bm{h}_i$ ($1 \leqslant i \leqslant t-1$) is using a matrix $\bm{W}_{\alpha} \in \mathbb{R}^{2p \times 2p}$, and calculating the weight as:
\begin{equation}\label{general_attention}
\alpha_{ti} = \bm{h}_t^{\top} \bm{W}_{\alpha}\bm{h}_i,
\end{equation}
and the context vector $\bm{c}_t$ can be obtained using Eq.~(\ref{local_attention}) and Eq.~(\ref{ct}).

\smallskip$\bullet$ \emph{Concatenation-based Attention}.
Another way to calculate the context vector $\bm{c}_t$ is using a multi-layer perceptron (MLP)~\cite{bahdanau2015neural}. We first concatenate the current hidden state $\bm{h}_s$ and the previous state $\bm{h}_i$, and then a latent vector can be obtained by multiplying a weight matrix $\bm{W}_{\alpha} \in \mathbb{R}^{q \times 4p}$, where $q$ is the latent dimensionality. We select $tanh$ as the activation function. The attention weight vector is generated as follows:
\begin{equation}\label{concatenation_attention}
\alpha_{ti} = \bm{v}_{\alpha}^{\top} \text{tanh}(\bm{W}_{\alpha}[\bm{h}_t; \bm{h}_i]),
\end{equation}
where $\bm{v}_{\alpha} \in \mathbb{R}^q$ is the parameter to be learned, and we can obtain the context vector $\bm{c}_t$ with Eq.~(\ref{local_attention}) and Eq.~(\ref{ct}).

\medskip\noindent\textbf{Diagnosis Prediction}\\
Given the context vector $\bm{c}_t$ and the current hidden state $\bm{h}_t$, we employ a simple concatenation layer to combine the information from both vectors to generate an attentional hidden state as follows:
\begin{equation}\label{attention_state}
\tilde{\bm{h}}_t = \text{tanh} (\bm{W}_c [\bm{c}_t; \bm{h}_t]),
\end{equation}
where $\bm{W}_c \in \mathbb{R}^{r \times 4p}$ is the weight matrix. The attentional vector $\tilde{\bm{h}}_t$ is fed through the softmax layer to produce the $(t+1)$-th visit information defined as:
\begin{equation}\label{eq:softmax}
\hat{\bm{y}}_{t} = \text{Softmax}(\bm{W}_s \tilde{\bm{h}}_t + \bm{b}_s)
\end{equation}
where $\bm{W}_s \in \mathbb{R}^{|\mathcal{G}| \times r}$ and $\bm{b}_s \in \mathbb{R}^{|\mathcal{G}|}$ are the parameters to be learned.

\medskip\noindent\textbf{Objective Function}\\
Based on Eq.~(\ref{eq:softmax}), we use the cross-entropy between the ground truth visit information $\bm{y}_{t}$ and the predicted visit $\hat{\bm{y}}_{t}$ to calculate the loss for all the patients as follows:
\begin{equation}
{\small
\begin{split}
&\mathcal{L}(\bm{x}_1^{(1)}, \cdots, \bm{x}_{T^{(1)}-1}^{(1)}, \cdots, \bm{x}_1^{(N)}, \cdots, \bm{x}_{T^{(N)}-1}^{(N)}) \\
= & - \frac{1}{N} \sum_{n=1}^N \frac{1}{T^{(n)}-1}
\sum_{t=1}^{T^{(n)} -1} \left( \bm{y}_{t}^{\top} \log(\hat{\bm{y}}_{t}) + (\bm{1} - \bm{y}_{t})^{\top} \log(\bm{1} - \hat{\bm{y}}_{t}) \right)
\end{split}
}
\end{equation}

\subsection{Interpretation}
In healthcare, the interpretability of the learned representations of medical codes and visits is important. We need to understand the clinical meaning of each dimension of medical code representations, and analyze which visits are crucial to the prediction.

Since the proposed model is based on attention mechanisms, it is easy to find the importance of each visit for prediction by analyzing the attention scores. For the $t$-th prediction, if the attention score $\alpha_{ti}$ is large, then the probability of the $(i+1)$-th visit information related to the current prediction is high.
We employ the simple method proposed in~\cite{choi2016mlr} to interpret the code representations.
We first use $\text{ReLU}(\bm{W}_v^{\top})$, a non-negative matrix, to represent the medical codes. Then we rank the codes by values in a reverse order for each dimension of the hidden state vector. Finally, the top $k$ codes with the largest values are selected as follows:
\begin{equation*}
\text{argsort}(\bm{W}_v^{\top}[:, i])[1:k],
\end{equation*}
where $\bm{W}_v^{\top}[:,i]$ represents the $i$-th column or dimension of $\bm{W}_v^{\top}$. By analyzing the selected medical codes, we can obtain the clinical interpretation of each dimension. 
Detailed examples and analysis are given in Section~\ref{4.4} and~\ref{4.5}.

%% file: 5-experiments.tex
\section{Experiments}\label{experiments}
In this section, we evaluate the performance of the proposed {\sf Dipole} model on two real world insurance claims datasets, compare its performance with other state-of-the-art prediction models, and show that it yields higher accuracy.

\subsection{Data Description}\label{data_description}
The datasets we used in the experiments are the Medicaid claims and the Diabetes claims.

\medskip\noindent\textbf{The Medicaid Dataset}\\
Our first dataset consists of Medicaid claims\footnote{\url{https://www.medicaid.gov}} over the year 2011. It consists of data corresponding to $147,810$ patients, and $1,055,011$ visits. The patient visits were grouped by week, and we excluded patients who made less than two visits.


\medskip\noindent\textbf{The Diabetes Dataset}\\
The Diabetes dataset consists of Medicaid claims over the years 2012 and 2013, corresponding to patients who have been diagnosed with diabetes (i.e. Medicaid members who have the ICD-9 diagnosis code 250.xx in their claims). It contains data corresponding to $22,820$ patients with $466,732$ visits.
The patient visits were aggregated by week, and excluded patients who made less than five visits.

\medskip
For both datasets, each visit information includes the ICD-9 diagnosis codes and  procedure codes, categorized in accordance with the Current Procedural Terminology (CPT). Table \ref{tab:dataset} lists more details about the two datasets.

\begin{table}[htb]
\centering
\caption{Statistics of Diabetes and Medicaid Dataset.}\label{tab:dataset}
\vspace{-0.15in}
\begin{tabular}{lcc}
\toprule
Dataset & Diabetes & Medicaid  \\ \midrule
\# of patients	& 22,820	&147,810\\
\# of visits	&466,732	&1,055,011\\
Avg. \# of visits per patient	&20.45	&7.14\\ \hline
\# of unique medical codes	&7,399	&8,522\\
- \# of unique diagnosis codes	&984	&1,021\\
- \# of unique procedure codes	&6,415	&7,501\\
Avg. \# of medical codes per visit	&6.35	&5.57\\
Max \# of medical codes per visit	&105	&99\\ \hline
\# of category codes	&422	&426\\
- \# of unique diagnosis categories	&183	&185\\
- \# of unique procedure categories	&239	&241\\
Avg. \# of category codes per visit	&4.85	&4.08\\
Max \# of category codes per visit	&38	&42\\
 \bottomrule
\end{tabular}
\end{table}

\subsection{Experimental Setup}
In this subsection, we first describe the state-of-the-art approaches for EHR representation learning and diagnosis prediction which are used as baselines, and then outline the measures used for evaluation. Finally, we introduce the implementation details.

\medskip\noindent\textbf{Baseline Approaches}\\
To validate the performance of the proposed model for diagnosis prediction task, we compare it with several state-of-the-art models.
We select three existing approaches as baselines\footnote{{\sf GRAM}~\cite{choi2016gram} is not a baseline as it uses external knowledge to learn the medical code representations.}:

{\sf Med2Vec}~\cite{choi2016mlr}. {\sf Med2Vec}, which follows the idea of Skip-gram \cite{mikolov2013distributed}, is a simple and robust algorithm to efficiently learn medical code representations and predict the medical codes appearing in the following visit based on the current visit information.

{\sf RETAIN}~\cite{choi2016retain}. {\sf RETAIN} is an interpretable predictive model in healthcare with reverse time attention mechanism, a two-level neural attention model. It can find influential past visits and important medical codes within those visits. Since the original {\sf RETAIN} is used for binary prediction task, we change the final softmax function for satisfying multiple variable prediction, i.e., diagnosis prediction.

{\sf RNN}. We first embed visit information into vector representations according to Eq.~(\ref{eq:visit}), then feed this embedding to the GRU. The hidden states produced by the GRU are directly used to predict the medical codes of the $(t+1)$-th visit using softmax according to Eq.~(\ref{eq:softmax}). All the parameters are trained together with the GRU.

\medskip\noindent\textbf{Our Approaches}\\
Since all the attention mechanisms proposed in Section \ref{sub_model} can be used for {\sf RNN} model, we propose three variants of {\sf RNN} as follows:

{\sf RNN$_l$}. We add location-based attention model into {\sf RNN}. The attention scores are calculated by Eq.~(\ref{location}). Then we can obtain the context vectors according to Eq.~(\ref{ct}). Based on the context vectors, we can generate attention hidden states using Eq.~(\ref{attention_state}). Finally, we can predict the medical codes of the $(t+1)$ visit using Eq.~(\ref{eq:softmax}).

{\sf RNN$_g$}. {\sf RNN$_g$} is similar to {\sf RNN$_l$}, but uses general attention model, i.e., Eq.~(\ref{general_attention}), to calculate attention scores.

{\sf RNN$_c$}. {\sf RNN$_c$} uses concatenation-based attention mechanism (Eq.~(\ref{concatenation_attention})) to calculate attention weights.

The proposed {\sf Dipole} model is a general framework for predicting diagnoses in healthcare. We show the performance of the following four approaches in the experiments.

{\sf Dipole$-$}. This model only uses the hidden states generated by BRNN to predict the next visit information, i.e., without employing any attention mechanisms.

{\sf Dipole$_l$}. It is based on location-based attention mechanism with Eq.~(\ref{location}).

{\sf Dipole$_g$}. {\sf Dipole$_g$} uses general attention model when calculating the context vectors, i.e., Eq.~(\ref{general_attention}).

{\sf Dipole$_c$}. Similar to {\sf Dipole$_l$} and {\sf Dipole$_g$}, {\sf Dipole$_c$} employs concatenation based attention mechanism (Eq.~(\ref{concatenation_attention})) in the predictive model.

\medskip\noindent\textbf{Evaluation Strategies}\\
To evaluate the performance of predicting future medical codes for each method, we use two measures: accuracy and accuracy$@k$ . Accuracy is defined as the correct medical codes ranked in top $k$ divided by $|\bm{y}_{t}|$, where $|\bm{y}_{t}|$ is the number of medical codes in the $(t+1)$-th visit, and $k$ equals to $|\bm{y}_{t}|$. Accuracy$@k$ is defined as the correct medical codes in top $k$ divided by $\min(k, |\bm{y}_{t}|)$. In our experiments, we vary $k$ from $5$ to $30$.


\medskip\noindent\textbf{Implementation Details}\\
We implement all the approaches with Theano 0.7.0~\cite{bergstra2010theano}. For training models, we use Adadelta~\cite{zeiler2012adadelta} with a mini-batch of 100 patients\footnote{For {\sf Med2Vec}, we use 1000 visits per batch as in~\cite{choi2016mlr}. The window size is 5 on the Diabetes dataset and 3 on the Medicaid dataset.}.
We randomly divide the dataset into the training, validation and testing set in a 0.75:0.1:0.15 ratio. The validation set is used to determine the best values of parameters. We also use regularization ($\l_2$ norm with the coefficient 0.001) and drop-out strategies (the drop-out rate is 0.5) for all the approaches. In the experiments, we set the same $m=256$, $p=256$ and $q=128$ for baselines and our approaches. We perform 100 iterations and report the best performance for each method.

\subsection{Results of Diagnosis Prediction}\label{results_diagnosis_prediction}
Table~\ref{tab:prediction} shows the accuracy of the proposed approaches and baselines on both Diabetes and Medicaid datasets for the diagnosis prediction task. $\# C$ represents the average number of correct predictions. The number of visits or predictions in the test set is  65,975 in the Diabetes dataset, and 136,023 in the Medicaid dataset.

\begin{table}[htb]
\centering
\newcommand{\tabincell}[2]{\begin{tabular}{@{}#1@{}}#2\end{tabular}}
\caption{The Accuracy of Diagnosis Prediction Task.}\label{tab:prediction}
\vspace{-0.15in}
{\small
\begin{tabular}{cccccc}
\toprule
\multicolumn{2}{c}{\multirow{2}{*}{Method}}& \multicolumn{2}{c}{Diabetes} & \multicolumn{2}{c}{Medicaid}\\
\cmidrule(r){3-4}\cmidrule(r){5-6}
                            && \# C & Accuracy & \# C  & Accuracy \\ \midrule
\multirow{3}{*}{Baseline}
&{\sf RNN}          &28,608                       & 0.4336      &  58,245                     &  0.4282 \\
&{\sf Med2Vec}   & 29,175                      & 0.4422      & 62,326                      & 0.4582 \\
&{\sf RETAIN}     & 28,851                     & 0.4373      & 63,496                     & 0.4668 \\ \hline
\multirow{7}{*}{\tabincell{c}{Our \\ Approach}}
&{\sf RNN$_l$}    & 29,880                     & 0.4529      &  62,938                     & 0.4627 \\
&{\sf RNN$_g$}   & 30,124                     & 0.4566      & 62,571                     & 0.4600\\
&{\sf RNN$_c$}   & 30,164                     & 0.4572      & 62,693                     & 0.4609 \\
&{\sf Dipole$-$}    & 29,623                     & 0.4490      & 62,475                     & 0.4593 \\
&{\sf Dipole$_l$}  & 30,645                     & 0.4645      & \textbf{65,441} & \textbf{0.4811} \\
&{\sf Dipole$_g$} & 29,464                     & 0.4466        & 64,557                     & 0.4746 \\
&{\sf Dipole$_c$} & \textbf{30,698} & \textbf{ 0.4653} & 63,931                     & 0.4700 \\
 \bottomrule
\end{tabular}
}
\end{table}

In Table~\ref{tab:prediction}, we can observe that the accuracy of the proposed approaches, including {\sf Dipole} and {\sf RNN} variants, is higher than that of baselines on the Diabetes dataset. Since most medical codes are about diabetes, {\sf Med2Vec} can correctly learn vector representations on the Diabetes dataset. Thus, {\sf Med2Vec} achieves the best results among the three baselines. For the Medicaid dataset, the accuracy of {\sf RETAIN} is better than that of {\sf Med2Vec}. The reason is that there are many diseases in the Medicaid dataset, and the categories of medical codes are more than those on the Diabetes dataset. In this case, attention mechanism can help {\sf RETAIN} to learn reasonable parameters and make correct prediction.

The accuracy of {\sf RNN} is the lowest among all the approaches on both datasets. This is because the prediction of {\sf RNN} mostly depends on the recent visits' information. It cannot memorize all  the past information. However, {\sf RETAIN} and the proposed {\sf RNN} variants, {\sf RNN$_l$}, {\sf RNN$_g$} and {\sf RNN$_c$}, can fully take all the previous visit information into consideration, assign different attention scores for past visits, and achieve better performance when compared to {\sf RNN}.

Since most of the visits on the Diabetes dataset are related to diabetes, it is easy to predict the medical codes in the next visit according to the past visit information. {\sf RETAIN} uses a reverse time attention mechanism for prediction, which will decrease the prediction performance compared with the approaches using a time ordered attention mechanism. Thus, the performance of the three proposed {\sf RNN} variants is better than that of {\sf RETAIN}. However, the accuracy of {\sf RETAIN} is better than the proposed {\sf RNN} variants' as the data are about different diseases on the Medicaid dataset. Using the reverse time attention mechanism can help us to learn the correct relationships among visits.

Both {\sf RNN} and the proposed {\sf Dipole$-$} do not use any attention mechanism, but the accuracy of {\sf Dipole$-$} is higher than that of {\sf RNN} on both the Diabetes and Medicaid dataset. It shows that modeling visit information from two directions can improve the prediction performance. Thus, it is reasonable to employ bidirectional recurrent neural networks for diagnosis prediction task.

The proposed {\sf Dipole$_c$} and {\sf Dipole$_l$} can achieve the best performance on the Diabetes and Medicaid dataset respectively, which shows that both modeling visits from two directions and assigning a different weight to each visit can improve the accuracy for diagnosis prediction task in healthcare. On the Diabetes dataset, {\sf Dipole$_l$} and {\sf Dipole$_c$} outperform all the baselines and the proposed {\sf RNN} variants. On the Medicaid dataset, the performance of all the three proposed approaches, {\sf Dipole$_l$}, {\sf Dipole$_g$} and {\sf Dipole$_c$} is better than that of baselines and {\sf RNN} variants.

Table~\ref{tab:acc_k} shows the experimental results with the accuracy$@k$ measurement on the Diabetes and Medicaid dataset separately. We can observe that as $k$ increases, the performance of all the approaches improves, but the proposed {\sf Dipole} approaches show superior predictive performance, demonstrating their applicability in predictive healthcare modeling. In Table~\ref{tab:acc_k}, {\sf RETAIN} can achieve comparable performance with the proposed approaches on the Medicaid dataset. The overall performance of location-based attention methods, {\sf Dipole$_l$} and {\sf RNN$_l$}, is better than that of other methods, which indicates that location-based attention performs well on this dataset. {\sf RETAIN} also uses location-based attention mechanism. Thus, it can obtain high accuracy.

\begin{table*}[htb]
\centering
\caption{The Accuracy$@k$ of Diagnosis Prediction Task.}\label{tab:acc_k}
\vspace{-0.15in}
\begin{tabular}{cccccccccccc}
\toprule
Dataset & Accuracy$@k$ & {\sf RNN} & {\sf Med2Vec} & {\sf RETAIN} & {\sf RNN$_l$} & {\sf RNN$_g$}& {\sf RNN$_c$} & {\sf Dipole$-$} & {\sf Dipole$_l$}& {\sf Dipole$_g$} & {\sf Dipole$_c$}\\\midrule
\multirow{6}{*}{Diabetes}
& 5  & 0.5236	&0.5210	&0.5257	&0.5389	&0.5423	&0.5418	&0.5413	&0.5568	&0.5350	&\textbf{0.5575} \\
&10 & 0.6107	&0.6015	&0.6102	&0.6281	&0.6316	&0.6317	&0.6325	&0.6466	&0.6204	&\textbf{0.6469} \\
&15 & 0.6829	&0.6744	&0.6826	&0.6993	&0.7026	&0.7034	&0.7036	&0.7146	&0.6913	&\textbf{0.7150} \\
&20 & 0.7355	&0.7257	&0.7354	&0.7515	&0.7542	&0.7555	&0.7548	&\textbf{0.7642}	&0.7415	&0.7641 \\
&25 & 0.7759	&0.7662	&0.7761	&0.7928	&0.7949	&0.7959	&0.7942	&\textbf{0.8021}	&0.7820	&0.8019 \\
&30 & 0.8087	&0.7998	&0.8091	&0.8259	&0.8273	&0.8279	&0.8251	&\textbf{0.8318}	&0.8141	&0.8316 \\ \hline
\multirow{6}{*}{Medicaid}
& 5   & 0.5238	& 0.5470	& 0.5663	& 0.5579	& 0.5540	& 0.5569	& 0.5586	& \textbf{0.5791}	& 0.5698	& 0.5660 \\
&10  & 0.6237	& 0.6342	& 0.6620	& 0.6645	& 0.6595	& 0.6627	& 0.6575	& \textbf{0.6764}	& 0.6663	& 0.6615 \\
&15  & 0.6933	& 0.7015	& 0.7297	& 0.7348	& 0.7300	& 0.7329	& 0.7249	& \textbf{0.7420}	& 0.7324	& 0.7268 \\
&20  & 0.7444	& 0.7511	& 0.7773	& 0.7842	& 0.7809	& 0.7828	& 0.7720	& \textbf{0.7877}	& 0.7785	& 0.7736 \\
&25  & 0.7843	& 0.7902	& 0.8134	& 0.8211	& 0.8183	& 0.8197	& 0.8074	& \textbf{0.8213}	& 0.8127	& 0.8088 \\
&30  & 0.8157	& 0.8211	& 0.8416	& \textbf{0.8496}	& 0.8469	& 0.8482	& 0.8358	& 0.8475	& 0.8400	& 0.8362 \\
 \bottomrule
\end{tabular}
\end{table*}

\subsection{Case Study}\label{4.4}
\begin{figure}[!htb]
 \centering
  \includegraphics[width=2.5in]{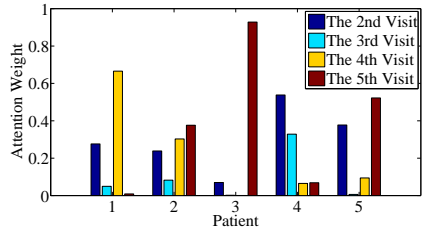}
  \vspace{-0.15in}
 \caption{Attention Mechanism Analysis.}\label{fig:attention}
\end{figure}

To demonstrate the benefit of applying attention mechanisms in diagnosis prediction task, we analyze the attention weights learned from one of the proposed approach {\sf Dipole$_c$}, which uses concatenation based attention mechanism. Figure~\ref{fig:attention} shows a case study for predicting the medical code in the sixth visit ($\bm{y}_5$) based on the previous visits on the Diabetes dataset. The concatenation-based attention weights are calculated for the visits from the second visit to the fifth visit according to the hidden states $\bm{h}_1$, $\bm{h}_2$, $\bm{h}_3$ and $\bm{h}_4$. Thus, we have four attention scores. In Figure~\ref{fig:attention}, X-axis represents patients, and Y-axis is the attention weight calculated for each visit. In this case study, we select five patients. We can observe that for different patients, the attention scores learned by the attention mechanism are different.

To illustrate the correctness of the learned attention weights, we provide an example. For the second patient in Figure~\ref{fig:attention}, we list all the diagnosis codes in Table~\ref{tab:case}. In order to predict the medical codes in the sixth visits, we first obtain the attention scores $\bm{\alpha} = [0.2386, 0.0824, 0.3028,0.3762]$. Analyzing this attention vector, we can conclude that the medical codes in the second, fourth and fifth visits significantly contribute to the final prediction. From Table~\ref{tab:case}, we can observe that the patient suffered \emph{essential hypertension} in the second and fourth visits, and diagnosed \emph{diabetes} in the fifth visits. Thus, the probability of the sixth visit's medical codes about \emph{diabetes} and diseases related to \emph{essential hypertension} is high. According to the proposed approach, we can predict the correct diagnoses that this patient suffers \emph{diabetes} and \emph{hypertensive heart disease}. This case study demonstrates that we can learn an accurate attention weight for each visit, and the experimental results in Section~\ref{results_diagnosis_prediction} also illustrate that the appropriate attention models can significantly improve the performance of the diagnosis prediction task in healthcare.

\begin{table}[tb]
\centering
\newcommand{\tabincell}[2]{\begin{tabular}{@{}#1@{}}#2\end{tabular}}
\caption{Diagnosis Codes in Each Visit for Patient 2 in the Case Study.}\label{tab:case}
\vspace{-0.15in}
{\small
\begin{tabular}{c|l}
\toprule
Visit& Diagnosis Codes \\ \midrule
1 & \tabincell{l}{Symptoms involving digestive system (787) \\ Essential hypertension (401)\\ Symptoms involving respiratory system and other\\ chest symptoms (786)\\ Special screening for other conditions (V82)} \\ \hline
2 & \tabincell{l}{Essential hypertension (401) }\\\hline
3 & \tabincell{l}{Disorders of lipoid metabolism (272) \\ Hypotension (458)}\\\hline
4 & \tabincell{l}{Essential hypertension (401) \\ Need for isolation and other prophylactic\\ measures (V07)}\\\hline
5 & \tabincell{l}{Diabetes mellitus (250)}\\\hline
6 & \tabincell{l}{Hypertensive heart disease (402) \\ Diabetes mellitus (250)}\\
 \bottomrule
\end{tabular}
}
\end{table}

\subsection{Code Representation Analysis}\label{4.5}
\begin{table*}[!htb]
\centering
\newcommand{\tabincell}[2]{\begin{tabular}{@{}#1@{}}#2\end{tabular}}
\caption{Interpretation for Diagnosis Code Representations on the Diabetes Dataset.}\label{tab:code}
\vspace{-0.15in}
{\footnotesize
\begin{tabular}{|p{160pt}<{\centering}|p{160pt}<{\centering}|p{160pt}<{\centering}|}
\hline
Dimension 10 & Dimension 38 & Dimension 77\\ \hline
\tabincell{l}{Glaucoma (365) \\ Fracture of one or more tarsal and\\ metatarsal bones (825) \\ Dementias (290) \\ Psoriasis and similar disorders (696) \\ Mild mental retardation (317) \\  Cataract (366) \\  Injury, other and unspecified (959) \\ Rheumatoid arthritis and other \\ inflammatory polyarthropathies (714) \\ Thyrotoxicosis with or without goiter(242) \\  Blindness and low vision (369)}
&
\tabincell{l}{Hereditary and idiopathic peripheral\\neuropathy (356) \\ Other disorders of soft tissues (729) \\ Dermatophytosis (110) \\ Other disorders of urethra and urinary \\track (599) \\ Mononeuritis of lower limb (355) \\ Diabetes mellitus (250) \\ Mononeuritis of upper limb \\and mononeuritis multiplex (354) \\ Sprains and strains of \\sacroiliac region (846) \\ Osteoarthrosis and allied disorders (715) \\ Other and unspecified disorders of back (724)}
&
\tabincell{l}{Cardiac dysrhythmias (427) \\ Chronic pulmonary heart disease (416) \\ Special screening for malignant \\neoplasms (V76) \\ Angina pectoris (413) \\ Other hernia of abdominal cavity without\\ mention of obstruction (553) \\ Cardiomyopathy (425) \\ Ill-defined descriptions and complications \\of heart disease (429) \\ Diabetes mellitus (250) \\  Acute pulmonary heart disease (415) \\ Gastrointestinal hemorrhage (578)}
\\ \hline
Dimension 79 & Dimension 141 & Dimension 142\\ \hline
\tabincell{l}{Neurotic disorders (300) \\ Other current conditions in the mother \\classifiable elsewhere (648) \\ Symptoms concerning nutrition metabolism \\and development (783) \\ Obesity and other hyperalimentation (278) \\ Diseases of esophagus (530) \\ Other organic psychotic conditions \\(chronic) (294) \\ Schizophrenic disorders (295) \\ Asthma (493) \\ Chronic liver disease and cirrhosis (571) \\ Spondylosis and allied disorders (721)}
&
\tabincell{l}{Viral hepatitis (070) \\ Other cellulitis and abscess (682) \\ Other personal history presenting \\hazards to health (V15) \\ Cellulitis and abscess of finger and toe (681) \\ Bacterial infection in conditions classified \\elsewhere (041) \\ Episodic mood disorders (296) \\ Chronic ulcer of skin (707) \\ Mononeuritis of upper limb and \\mononeuritis multiplex (354) \\ Other diseases due to viruses and\\ Chlamydiae (078) \\ Diabetes mellitus (250)}
&
\tabincell{l}{Essential hypertension (401) \\ Hypertensive renal disease (403) \\ Hypertensive heart disease (402) \\ Chronic renal failure (585) \\ Other disorders of kidney and ureter (593) \\ Other psychosocial circumstances (V62) \\ Secondary hypertension (405) \\ Nonspecific abnormal results of \\function studies (794) \\ Calculus of kidney and ureter (592) \\ Other organic psychotic conditions \\(chronic) (294)}
 \\ \hline
\end{tabular}
}
\end{table*}

The interpretability of medical codes is important in healthcare. In order to analyze the representations of medical codes learned by the proposed model {\sf Dipole$_g$}, we show top ten diagnosis codes with the largest value in each of six columns selected from the hidden representation matrix $\bm{W}_v^{\top} \in \mathbb{R}^{|\mathcal{C}| \times m}$ in Table~\ref{tab:code}. In this way, we can demonstrate the characteristic of each column and map each dimension from the code embedding  space to the medical concept.

In Table~\ref{tab:code}, we can clearly observe that the codes in all the six dimensions are about diabetes complications, which are in accordance with the complications listed on the American Diabetes Association\footnote{\url{http://www.diabetes.org/living-with-diabetes/complications/}}. Dimension 10 is related to eye complications and Alzheimer's disease. Diabetes can damage the blood vessels of the retina (diabetic retinopathy), potentially leading to blindness, and Type 2 diabetes may increase the risk of Alzheimer's disease. Dimension 38 relates to the complications of neuropathy (nerve damage). Dimension 77 is about heart diseases. It has been established that there is a high correlation between diabetes, heart disease, and stroke. In fact, two out of three patients with diabetes die from heart disease or stroke. Patients with diabetes have a greater risk of depression than people without diabetes. Dimension 79 includes the codes related to mental health. Dimension 141 shows a fact that diabetes may cause skin problems, including bacterial and fungal infections. High blood pressure is one common complication of diabetes shown in dimension 142, which also raises the risk for heart attack, stroke, eye problems, and kidney disease.

\subsection{Assumption Validation}
\begin{figure}
  \centering
  \subfigure[Diabetes Dataset]{
    \label{fig:subfig:a} 
    \includegraphics[width=1.8in]{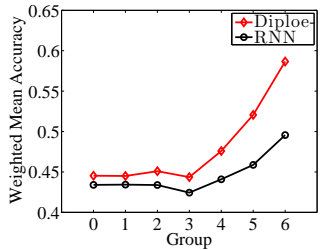}}
  \subfigure[Medicaid Dataset]{
    \label{fig:subfig:b} 
    \includegraphics[width=1.8in]{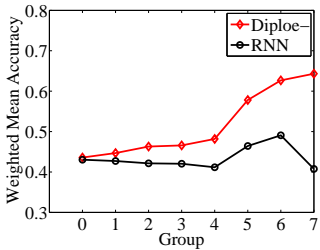}}
    \vspace{-0.15in}
  \caption{Weighted Mean Accuracy of Different Groups.}
  \label{fig:assu} 
\end{figure}

In the proposed model, we adopt bidirectional recurrent neural networks to model patient visits instead of recurrent neural networks. To illustrate the benefit of employing bidirectional recurrent neural networks, we analyze the detailed mean accuracy of {\sf RNN} and {\sf Dipole-} shown in Figure \ref{fig:assu}. We first divide patients into different groups based on the number of visits. The group label is the quotient of the number of visits divided by 15 for the Diabetes dataset and 7 for the Medicaid dataset, which is X-axis in Figure \ref{fig:assu}. Then we calculate the weighted average accuracy (Y-axis) of different groups, i.e., $\frac{\sum_{n} {MA}_n * C_n}{\sum_{n} C_n}$, where ${MA}_n$ is the mean accuracy of all the patients with $n$ visits, and $C_n$ is the number of patients with $n$ visits. From Figure \ref{fig:assu}, we can observe that the average accuracy of {\sf Dipole-} is better than that of {\sf RNN} in different groups. On the Diabetes dataset, the weighted mean accuracy of {\sf RNN} increases when the number of visits becomes larger. This is because the codes of visits on the Diabetes dataset are all about diabetes, and {\sf RNN} can make correct prediction according to recent visits' information. However, the codes on the Medicaid dataset are related to multiple diseases, and it is hard to correctly predict the future visit information when the sequences are too long. Thus, the weighted mean accuracy significantly drops when the number of visits is large on the Medicaid dataset.

\begin{figure}
  \centering
  \subfigure[Diabetes Dataset]{
    \label{fig:subfig:a} 
    \includegraphics[width=1.8in]{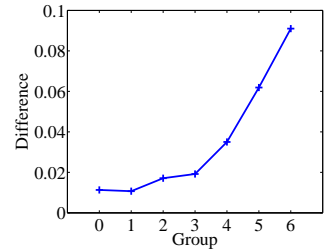}}
  \subfigure[Medicaid Dataset]{
    \label{fig:subfig:b} 
    \includegraphics[width=1.8in]{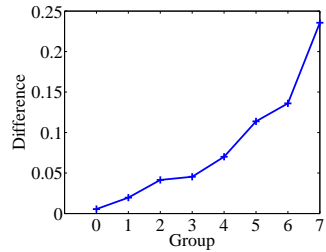}}
    \vspace{-0.15in}
  \caption{Difference of Weighted Mean Accuracy.}
  \label{fig:assu_diff} 
\end{figure}

Figure \ref{fig:assu_diff} shows the difference of weighted mean accuracy between {\sf Dipole-} and {\sf RNN} in different groups. We can observe that with the increase of the number of visits, the difference also augments dramatically. It demonstrates that bidirectional recurrent neural networks can ``remember'' more information when the sequences are long, and make correct predictions with their memories. Thus, modeling patient visits with bidirectional recurrent neural networks is reasonable.
\vspace{-0.1in}

%% file: 7-conclusions.tex
\section{Conclusions}\label{end}
Diagnosis prediction is a challenging and important task, and interpreting the prediction results is a hard and vital problem for predictive model in healthcare.
Many existing work in diagnosis prediction employs deep learning techniques, such as recurrent neural networks (RNNs), to model the temporal and high dimensional EHR data. 
However, RNN-based approaches may not fully remember all the previous visit information, which leads to the incorrect prediction.
To interpret the predicting results, existing work introduces location-based attention model, but this mechanism ignores the relationships between the current visit and the past visits.

In this paper, we propose a novel model, named {\sf Dipole}, to address the challenges of modeling EHR data and interpreting the prediction results.
By employing bidirectional recurrent neural networks, {\sf Dipole} can remember the hidden knowledge learned from the previous and future visits.
Three attention mechanisms allow us to interpret the prediction results reasonably.
Experimental results on two large real world EHR datasets prove the effectiveness of the proposed {\sf Dipole} for diagnosis prediction task.
Analysis shows that the attention mechanisms can assign different weights to previous visits when predicting the future visit information.
We demonstrate that the learned representations of medical codes are meaningful. 
Finally, an experiment is conducted to validate the reasonableness and effectiveness of modeling patient visits with bidirectional recurrent neural networks.